\documentclass{article}
\usepackage{a4wide}
\usepackage{times}
\usepackage{helvet}
\usepackage{courier}
\usepackage{amssymb,amsmath}
\usepackage{comment}
\usepackage{graphicx}

\newcommand{\ignore}[1]{}

\begin{document}
%
\title{Peek Arc Consistency}
\author{Manuel Bodirsky
\and Hubie Chen}


\date{ }

\maketitle

\begin{abstract}
\begin{quote}
This paper studies peek arc consistency, 
 a reasoning technique
that extends the well-known arc consistency technique
for constraint satisfaction.
In contrast to other more costly extensions of arc consistency that have
 been studied in the literature, peek arc consistency requires only
 linear space and quadratic time
 and can be parallelized in a straightforward way such that
 it runs in linear time with a linear number of processors.
We demonstrate that for various constraint languages,
peek arc consistency gives a polynomial-time decision procedure
for the constraint satisfaction problem.
We also present an
algebraic characterization of those constraint languages
that can be solved by peek arc consistency,
and study the robustness of the algorithm.
\end{quote}
\end{abstract}

\newtheorem{theorem}                            {Theorem}
\newtheorem{lemma}              [theorem]       {Lemma}
\newtheorem{remark}             [theorem]       {Remark}
\newtheorem{fact}               [theorem]       {Fact}
\newtheorem{claim}              [theorem]       {Claim}
\newtheorem{corollary}          [theorem]       {Corollary}
\newtheorem{definition}         [theorem]       {Definition}
\newtheorem{prop}               [theorem]       {Proposition}
\newtheorem{warning}            [theorem]       {Warning}
\newtheorem{observation}        [theorem]       {Observation}
\newtheorem{example}        [theorem]       {Example}

\newenvironment{pf}{\noindent\textbf{Proof\/}.}{$\Box$ \vspace{1mm}}

\newcommand{\rats}{\mathbb{Q}}

\newcommand{\rela}{\mathbf{A}}
\newcommand{\relb}{\mathbf{B}}
\newcommand{\relc}{\mathbf{C}}
\newcommand{\reld}{\mathbf{D}}
\newcommand{\relg}{\mathbf{G}}
\newcommand{\pow}{\wp}

\newcommand{\csp}{{\text{CSP}}}

\newcommand{\fancya}{\mathcal{A}}

\newcommand{\ind}{\mathrm{Ind}}

\newcommand{\homo}{\rightarrow}

\newcommand{\tup}[1]{\overline{#1}}

\section{Introduction}

\paragraph{Background.}
A basic knowledge reasoning task that has been studied
in many incarnations is to decide the satisfiability of 
given relationships on variables, 
where, for instance, variables may represent  objects
such as temporal events or spatial regions, and 
relationships may express precedence, containment, overlap, disjointness, and so forth.
Instances of this reasoning task can typically be modeled 
using the \emph{constraint satisfaction problem (CSP)}, 
a computational problem in which the input
consists of a set of constraints on variables, and the question is
whether or not there is an assignment to the variables satisfying all
of the constraints.
While the CSP is in general NP-hard, researchers have, in numerous settings,
aimed to identify restricted sets of relationships under which the CSP is
polynomial-time decidable; we refer to sets of relationships as
\emph{constraint languages}.

\emph{Arc consistency} is an algorithmic technique for constraint satisfaction
that has been heavily studied and for which highly efficient implementations
that are linear in both time and space are known. 
Arc consistency provides a one-sided satisfiability check.
It may detect an inconsistency, which always implies that the input
instance is unsatisfiable. While the converse does not hold
in general, it has been shown to hold for some particular constraint languages,
that is, arc consistency provides a decision procedure for satisfiability
for these languages. Examples include the language of boolean Horn clauses;
various graph homomorphism problems, for example,
homomorphisms to orientations of finite paths~\cite{HNBook};
and all constraint languages where satisfiability is 
first-order definable~\cite{AtseriasLICS05}.

Curiously, arc consistency typically 
cannot be used as a decision procedure 
for \emph{infinite-domain} constraint languages, by which we mean constraint
languages under which variables can take on infinitely many values.
In many cases, a reason for this is that arc consistency performs
inference by considering unary (arity $1$) projections of relations,
and all such projections are already equal to the full domain of the language.
As an example, consider the binary relations $\leq$ and $\neq$ interpreted
over the domain 
of rational numbers $\rats$.  For each of these relations, both of the
two possible unary projections are equal to $\rats$, and arc consistency
in fact will not perform any inference.

\emph{Path consistency} is a more powerful algorithmic technique that
 provides a polynomial-time decision procedure
for further finite and infinite domain constraint languages. However, the greater power
comes at the price of 
worse time and space complexity: the best known implementations
require cubic space and quadratic time.
 Unfourtunately, this makes the path consistency procedure
prohibitive for many applications where one has to deal with 
large instances of the constraint satisfaction problem.

\paragraph{Peek arc consistency.}
In this paper, we study a general algorithmic technique for constraint
satisfaction that we call \emph{peek arc consistency}.
Here, we describe the idea of the algorithm for finite-domain constraint
satisfaction, although, as we show in the paper,
this algorithm can be effectively applied to infinite-domain
constraint satisfaction as well.
The algorithm performs the following.
For each variable-value pair $(x,a)$, 
the variable $x$ is set to the value $a$, 
and then the arc consistency procedure is run
on the resulting instance of the CSP. If there is a variable
$x$ such that for all values $a$ the arc consistency procedure 
detects an inconsistency on $(x, a)$, then the algorithm reports
an inconsistency.
As with arc consistency, this algorithm provides a one-sided
satisfiability check.
One might concieve of this algorithm as being a step more sophisticated
than arc consistency; 
it invokes arc consistency as it takes a ``peek'' at each variable.

Peek arc consistency has many practical and theoretical selling points.
Arc consistency can be implemented in linear space, for any fixed
finite-domain constraint language and many infinite-domain constraint languages; the same holds for peek arc consistency.
The time complexity of peek arc consistency
is quadratic in the input size, which is still much 
better than the path-consistency algorithm, where the best
known implementations have a running time that is cubic in the
input size. 
Moreover, peek arc consistency can be parallelized
in a straightforward way: for each variable-value pair, the arc consistency
procedure can be performed on a different processor. Hence,
with a linear number of processors,
we achieve a linear running time, for a fixed constraint language.
We would also like to remark
that implementing peek arc consistency is straightforward
if one has access to an implementation of arc consistency as a subroutine.

We demonstrate that 
the class of constraint languages solvable by peek arc consistency
is a considerable extension
of that which can be solved by arc consistency, and in particular
contains many infinite-domain constraint languages. Examples
are the constraint satisfaction problem for the point algebra in temporal reasoning~\cite{PointAlgebra},
and tractable set constraints~\cite{DrakengrenJonssonSets}.
But also, several finite-domain constraint languages 
where previously the ``best''
known algorithm was the path-consistency procedure can be solved by our peek arc consistency
procedure. For example, this is the case for homomorphism problems to unbalanced orientations of cycles~\cite{FederCycles}.
Other examples that can be solved by peek arc consistency but not
by arc consistency are 2-SAT, and many other CSPs where the relations are closed under a dual descriminator or a median operation. 

Our study of peek arc consistency employs universal algebraic techniques
which have recently come into focus in the complexity of constraint
satisfaction.
In addition to obtaining results showing that languages are tractable
by this algorithm, we develop an algebraic characterization of
the constraint languages solvable by the algorithm.
The characterization is exact--necessary and 
sufficient--for all finite and infinite domain constraint languages.
We also exhibit closure properties on the class of constraint languages
tractable by the algorithm.

A notable 
feature of this work is the end to which universal algebraic
techniques are applied.
Thus far, in constraint satisfaction, such techniques
 have primarily been used to demonstrate complexity class inclusion 
results, such as polynomial-time decidability results,
and completeness results, such as NP-completeness results.
Here, we utilize such techniques to investigate the power of a
\emph{particular} efficient and practical algorithm.
That is, we differentiate among constraint languages 
depending on whether or not they are solvable via a specific
algorithmic method, as opposed to whether or not they are
contained in a complexity class.
To our knowledge, 
this attitude has only been adopted in a limited number of
previous papers that studied
arc consistency and extensions thereof~\cite{DalmauPearson,slaac}.

\ignore{
\paragraph{Related Work.}
Other variations and extensions of arc consistency have been studied
in the literature. \emph{Look-ahead arc consistency (LAAC)}
and \emph{Smart look-ahead arc consistency (SLAAC)} 
can also be implemented easily given an implementation of the
arc consistency procedure. Both
algorithms also give a one-sided test, but on the opposite side than
arc consistency and peek arc consistency: if LAAC or SLAAC  do not reject an instance, we can be sure that there is a solution
(without assumptions on the constraint language).  
Both LAAC and SLAAC cannot be parallelized in a similar
straightforward way as PAC. }

\section{Preliminaries}
Our definitions and notation are fairly standard.

\paragraph{Structures.}
A \emph{tuple} over a set $B$ is an element of $B^k$
for a value $k \geq 1$ called the \emph{arity} of the tuple;
when $\tup{t}$ is a tuple, we use the notation $\tup{t} = (t_1, \ldots, t_k)$
to denote its entries.
A \emph{relation} over a set $B$ is a subset of $B^k$ 
for a value $k \geq 1$ called the \emph{arity} of the relation.
A \emph{signature} $\sigma$ is a finite set of symbols, each of which
has an associated arity.  
We use $\pi_i$ to denote the operator that projects onto the $i$th coordinate:
$\pi_i(\tup{t})$ denotes the $i$th entry $t_i$ of a tuple
$\tup{t} = (t_1, \ldots, t_k)$, and for a relation $R$ we define
$\pi_i(R) = \{ \pi_i(\tup{t}) ~|~ \tup{t} \in R \}$.

A \emph{structure} $\relb$ over signature $\sigma$
consists of a universe $B$, which is a set,
and a relation $R^{\relb} \subseteq B^k$ for each symbol $R$ of arity $k$.
(Note that in this paper, we are concerned only with relational structures,
which we refer to simply as structures.)
Throughout, we will use the bold capital letters $\rela, \relb, \ldots$
to denote structures, and the corresponding non-bold capital letters
$A, B, \ldots$ to denote their universes.
We say that a structure $\relb$ is \emph{finite} if its universe
$B$ has finite size.

For two structures $\rela$ and $\relb$ over the same signature $\sigma$,
the product structure $\rela \times \relb$ is defined to be
the structure with universe $A \times B$ and such that
$R^{\rela \times \relb} = 
\{ ((a_1, b_1), \ldots, (a_k, b_k)) ~|~ \tup{a} \in R^{\rela}, \tup{b} \in R^{\relb} \}$ for all $R \in \sigma$.
We use $\rela^n$ to denote the $n$-fold product 
$\rela \times \cdots \times \rela$.

We say that a structure $\relb$ 
over signature $\sigma'$
is an \emph{expansion} of another
structure $\rela$ over signature $\sigma$
if (1) $\sigma' \supseteq \sigma$,
(2) the universe of $\relb$ is equal to the universe of $\rela$,
and 
(3) for every symbol $R \in \sigma$, it holds that
$R^{\relb} = R^{\rela}$.
We will use the following non-standard notation.
For any structure $\rela$ (over signature $\sigma$)
and any subset $S \subseteq A$,
we define $[\rela, S]$ to be the structure
with the signature
$\sigma \cup \{ U \}$
where $U$ is a new symbol of arity $1$, defined by
$U^{[\rela, S]} = S$ and $R^{[\rela, S]} = R^{\rela}$ for all $R \in \sigma$.

For two structures $\rela$ and $\relb$ over the same signature $\sigma$,
we say that $\rela$
is an \emph{induced substructure}
of $\relb$ if $A \subseteq B$
and for every $R \in \sigma$ of arity $k$, it holds that
$R^{\rela} = A^k \cap R^{\relb}$.
Observe that for a structure $\relb$ and a subset $B' \subseteq B$,
there is exactly one induced substructure of $\relb$ with universe $B'$.

\paragraph{Homomorphisms and the constraint satisfaction problem.}
For structures $\rela$ and $\relb$ over the same signature $\sigma$,
a \emph{homomorphism} from $\rela$ to $\relb$
is a mapping $h: A \rightarrow B$ such that
for every symbol $R$ of $\sigma$ and every tuple
$(a_1, \ldots, a_k) \in R^{\rela}$, it holds that
$(h(a_1), \ldots, h(a_k)) \in R^{\relb}$.
We use $\rela \homo \relb$ to indicate that there is a homomorphism
from $\rela$ to $\relb$; 
when this holds, we also say that $\rela$ \emph{is homomorphic to} $\relb$.
The homomorphism relation $\homo$ is transitive, that is,
if $\rela \homo \relb$ and $\relb \homo \relc$, then $\rela \homo \relc$.

For any structure $\relb$ (over $\sigma$), 
the \emph{constraint satisfaction problem for $\relb$}, denoted by $\csp(\relb)$, is the problem of
deciding, given as input a finite structure $\rela$ over $\sigma$,
whether or not
there exists a homomorphism from $\rela$ to $\relb$.
In discussing a problem of the form 
$\csp(\relb)$, we will refer to
 $\relb$ as the \emph{constraint language}.

There are several equivalent definitions of the constraint satisfaction
problem for a constraint language, most notably the definition
used in artificial intelligence. In logic, the constraint satisfaction problem can be formulated as the satisfiability problem for primitive positive formulas in a fixed structure $\relb$.
Homomorphism problems as defined above have been studied independently from artificial intelligence in graph theory, and the connection to constraint satisfaction problems has been observed in~\cite{FederVardi}.

\paragraph{pp-definability.}  
Let $\sigma$ be a signature; a \emph{primitive positive formula over $\sigma$}
is a formula
built from atomic formulas $R(w_1, \ldots, w_n)$ with $R \in \sigma$,
conjunction, and existential quantification.
A relation $R \subseteq B^k$ is
\emph{primitive positive definable (pp-definable)} in a structure $\relb$
(over $\sigma$) if there exists a primitive positive formula
$\phi(v_1, \ldots, v_k)$ with free variables $v_1, \ldots, v_k$
such that
$$(b_1, \ldots, b_k) \in R \Leftrightarrow \relb, b_1, \ldots, b_k \models \phi.$$

\paragraph{Automorphisms.}
An isomorphism between two relational structures $\rela$ and
$\relb$ over the same signature $\sigma$ is a bijective mapping from $A$ to $B$ such that 
$\tup t \in R^A$ if and only if $f(\tup t) \in R^B$ for all relation symbols
$R$ in $\sigma$. An automorphism of $\rela$ is an isomorphism between $\rela$ and $\rela$. An \emph{orbit} of $\rela$ is an
equivalence class of the equivalence relation $\equiv$ that is defined 
on $A$ by $x \equiv y$ iff $\alpha(x) = y$ for some automorphism $\alpha$ of $\rela$.

\paragraph{Polymorphisms.}
When $f: B^n \rightarrow B$ is an operation on $B$
and 
$\tup{t_1} = (t_{11}, \ldots, t_{1k}), \ldots, \tup{t_n} = (t_{n1}, \ldots, t_{nk}) \in B^k$ 
are tuples of the same arity $k$ 
over $B$, we use
$f(\tup{t_1}, \ldots, \tup{t_n})$ to denote the arity $k$ tuple
obtained by applying $f$ coordinatewise, that is, 
$f(\tup{t_1}, \ldots, \tup{t_n}) = 
(f(t_{11}, \ldots, t_{n1}), \ldots, f(t_{1k}, \ldots, t_{nk}))$.
An operation $f: B^n \rightarrow B$ is a \emph{polymorphism} 
of a structure $\relb$ over $\sigma$ if for every symbol $R \in \sigma$
and any tuples $\tup{t_1}, \ldots, \tup{t_n} \in R^{\relb}$, 
it holds that $f(\tup{t_1}, \ldots, \tup{t_n}) \in R^{\relb}$.
That is, each relation $R^{\relb}$ is closed under the action of $f$.
Equivalently, an operation $f: B^n \rightarrow B$ 
is a polymorphism of $\relb$ if
it is a homomorphism from $\relb^n$ to $\relb$.
Note that every automorphism is a unary polymorphism.

\paragraph{Categoricity.}
Several of our examples for constraint languages
over infinite domains will have the following property that
is of central importance in model theory. 
A countable structure is \emph{$\omega$-categorical} if all countable models
of its first-order theory\footnote{The \emph{first-order theory}
of a structure is the set of first-order sentences that is
true in the structure.} are isomorphic. By the Theorem of 
Ryll-Nardzewski (see e.g.~\cite{Hodges}) this is equivalent
to the property that for each $n$ there is a finite number of 
inequivalent first-order formulas over $\Gamma$ with $n$ 
free variables. A well-known example of an $\omega$-categorical
structure is $(\mathbb Q,<)$; for many more examples
of $\omega$-categorical structures
and their application to formulate well-known constraint
satisfaction problems, see~\cite{BodirskySurvey}.

\section{Arc Consistency}
\label{sect:ac}
In this section, we introduce the notion of arc consistency
that we will use, and review some related notions and results.
The definitions
we give 
apply to structures with relations of any arity, and not just
binary relations.  The notion 
of arc consistency studied here is sometimes called
\emph{hyperarc consistency}.
Our discussion is based on the paper~\cite{DalmauPearson}.

For a set $B$, let $\pow(B)$ denote the power set of $B$.
For a structure $\relb$ (over $\sigma$),
we define
$\pow(\relb)$ to be the structure with universe
$\pow(B) \setminus \{ \emptyset \}$ and where,
for every symbol $R \in \sigma$ of arity $k$,
$R^{\pow(\relb)} = \{ (\pi_1 S, \ldots, \pi_k S) ~|~ S \subseteq R^{\relb}, S \neq \emptyset \}$.

\begin{definition}
\label{def:acc}
An instance $\rela$ of  CSP$(\relb)$ has the 
\emph{arc consistency condition (ACC)} if
there exists a homomorphism from $\rela$ to $\pow(\relb)$.
\end{definition}

\begin{definition}
\label{def:ac-decides}
We say that arc consistency (AC) \emph{decides} $\csp(\relb)$
if for all finite structures $\rela$, the following holds:
$(\rela, \relb)$ has the ACC implies that 
$\rela \homo \relb$.
\end{definition}

Note that the converse of the condition given in this definition
always holds. 
By a \emph{singleton}, we mean a set containing exactly one element.

\begin{prop}
\label{prop:singleton-homomorphism}
For any structures $\rela$ and $\relb$, if 
$h$ is a homomorphism from $\rela$ to $\relb$, 
then the mapping that takes $a$ to the singleton $\{ h(a) \}$
is a homomorphism from $\rela$ to $\pow(\relb)$.
\end{prop}

Hence, when AC decides $\csp(\relb)$, an instance $\rela$ of 
$\csp(\relb)$ is a ``yes'' instance if and only if 
$\rela$ has the ACC with respect to $\relb$.  That is, deciding whether
an instance $\rela$ is a ``yes'' instance can be done just by checking
the ACC.
It was observed in~\cite{FederVardi} that, for any finite
structure $\relb$, there is an algebraic characterization of
AC: 
AC decides $\csp(\relb)$ if and only if there is a homomorphism
from $\pow(\relb)$ to $\relb$.

It is well-known that for a finite structure $\relb$,
whether or not instances $\rela$ of $\csp(\relb)$ have the ACC
can be checked
in polynomial-time. 
The algorithm for this is called the \emph{arc consistency procedure}, and it can be implemented in 
linear time and linear space in the size of $\rela$;
note that we consider $\relb$ to be fixed. 
The same holds for many infinite-domain
constraint languages, for example for all $\omega$-categorical constraint languages.
Since this is less well-known, and requires
a slightly less standard formulation of the arc consistency procedure,
we present a formal description of the algorithm that
we  use;  this algorithm can be applied for any (finite- and infinite-domain) constraint
language that has finitely
many pp-definable unary relations in $\relb$.

We assume that $\relb$ contains a relation for each unary primitive
positive definable relation in $\relb$. 
This is not a strong assumption, since we might
always study the expansion $\relb'$ of 
$\relb$ by all such unary relations. Then, if we are given
an instance $\rela$ of CSP$(\relb)$, we might run the algorithm
for CSP$(\relb')$ on the expansion $\rela'$ of $\rela$ that has the
same signature as $\relb'$ and where the new unary relations are interpreted by empty relations.
It is clear that a mapping from $A$ to $B$ is a homomorphism
from $\rela$ to $\relb$ if and only if it is a homomorphism from
$\rela'$ to $\relb'$.

\begin{figure*}
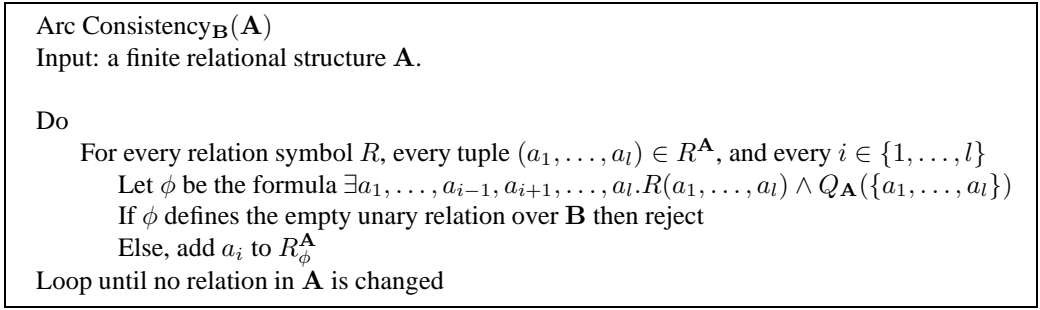

\begin{center}
\fbox{
\begin{tabular}{l}
Arc Consistency$_\relb(\rela)$ \\
Input: a finite relational structure $\rela$. \\
\\
Do \\
\hspace{0.5cm} For every relation symbol $R$,
every tuple $(a_1,\dots,a_l) \in R^\rela$, and every $i \in \{1,\dots,l\}$ \\
\hspace{1cm} Let $\phi$ be the formula $\exists a_1,\dots,a_{i-1},a_{i+1},\dots,a_l. R(a_1,\dots,a_l) \wedge Q_\rela(\{a_1,\dots,a_l\})$ \\
\hspace{1cm} If $\phi$ defines the empty unary relation over $\relb$ then reject \\
\hspace{1cm} Else, add $a_i$ to $R_\phi^\rela$ \\
Loop until no relation in $\rela$ is changed \\
\end{tabular}}
\end{center}
\caption{The arc consistency procedure for CSP$(\relb)$, where $\relb$ contains all primitive positive definable unary relations in $\relb$.}
\label{fig:ac}
\end{figure*}

To conveniently formulate the algorithm, we write
$R_{\phi(x)}$ for the relation symbol of the relation that is defined by 
a pp-formula $\phi(x)$ in $\relb$. 
We write $Q_\rela(\{a_1,\dots,a_l\})$ for the
conjunction over all formulas of the form $S(a_i)$ where $S$ is a unary relation symbol such that $a_i \in S^\rela$.
The pseudo-code of the arc consistency procedure 
can be found in Figure~\ref{fig:ac}.

The space requirements of the given arc consistency procedure are clearly linear. It is also well-known and easy to see that the procedure  can be implemented such that its running
time is \emph{linear} in the size of the input. 

\begin{prop}\label{prop:acc}
Let $\relb$ be a structure with finitely many 
primitive-positive definable unary relations. Then a given instance
$\rela$ of CSP$(\relb)$
has the ACC if and only if the arc consistency procedure presented
in Figure~\ref{fig:ac} does not reject.
\end{prop}

In particular, we can apply the algorithm shown in Figure~\ref{fig:ac}
to all constraint satisfaction problems with an $\omega$-categorical constraint language. 
However, it was shown that in this case 
the algorithm cannot be used as a decision procedure
for CSP$(\relb)$ (i.e., that rejects an instance $\rela$ if and only if it does not homomorphically map to $\relb$), 
unless $\relb$ is homomorphically equivalent 
to a finite structure~\cite{BodDal}.

\section{Peek Arc Consistency} \label{sect:pac}
We present basic definitions and results concerning
peek arc consistency.  The following two definitions
are analogous to 
Definitions~\ref{def:acc} and ~\ref{def:ac-decides}
of the previous section.

\begin{definition}
An instance $(\rela, \relb)$ of the CSP has the 
\emph{peek arc consistency condition (PACC)} if
for every element $a \in A$,
there exists a homomorphism $h$ from $\rela$ to $\pow(\relb)$
such that $h(a)$ is a singleton.
\end{definition}

\begin{definition}
We say that peek arc consistency (PAC) \emph{decides} $\csp(\relb)$
if for all finite structures $\rela$, the following holds:
$(\rela, \relb)$ has the PACC implies that $\rela \homo \relb$.
\end{definition}

The converse of the condition given in this definition always holds.
Suppose that $\rela \homo \relb$; then,
the mapping taking each $a \in A$ to the singleton $\{ h(a) \}$
is a homomorphism from $\rela$ to $\pow(\relb)$
(Proposition~\ref{prop:singleton-homomorphism}),
and hence $(\rela, \relb)$ has the PACC.

We now present an algorithm that decides for a given instance $\rela$
of CSP$(\relb)$, whether $(\rela, \relb)$ has the PACC. 
We assume that $\relb$ has a finite number of orbits and 
pp-definable binary relations. This holds in particular
for all $\omega$-categorical structures.
The following lemma then allows us to use the arc consistency procedure presented in Figure~\ref{fig:ac} for every expansion of 
$\relb$ by singletons.

\begin{lemma}\label{lem:fin}
Let $\relb$ be a structure with finitely many pp-definable binary relations. Then every expansion of $\relb$
by a constant has finitely
many pp-definable unary relations.
\end{lemma}
\begin{pf}
Suppose for contradiction that for a constant $b$, there are infinitely
many pairwise distinct unary relations
 with a pp-definition in the expansion of
 $\relb$ with $\{b\}$.
 For each such definition,
if we replace the occurrences
 of the relation symbol for the singleton $\{b\}$ by a new variable, 
we obtain formulas that are pp-definitions in $\relb$ of pairwise distinct binary relations.
\end{pf}

\begin{figure*}
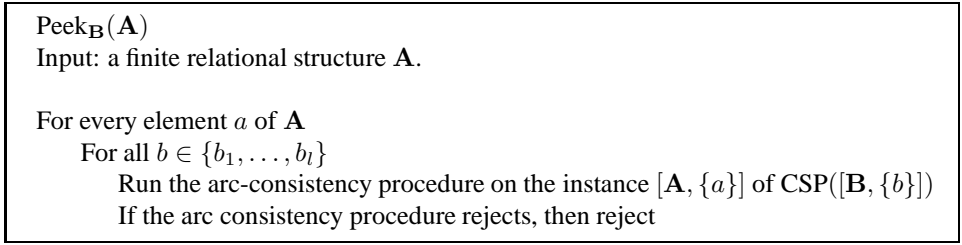

\begin{center}
\fbox{
\begin{tabular}{l}
Peek$_\relb(\rela)$ \\
Input: a finite relational structure $\rela$. \\
\\
For every element $a$ of $\rela$ \\
\hspace{0.5cm} For all $b \in \{b_1,\dots,b_l\}$ \\
\hspace{1cm} Run the arc-consistency procedure on the instance $[\rela,\{a\}]$ of CSP$([\relb,\{b\}])$ \\
\hspace{0.5cm} If the arc consistency procedure rejects for pp-definable sets, reject \\
\end{tabular}}
\end{center}
\caption{The peek arc consistency procedure for CSP$(\relb)$ for structures $\relb$ that have finitely many orbits and pp-definable binary relations. Let $b_1,\dots,b_l$ be arbitrary representatives from the orbits in $\relb$, i.e., we assume that there are $l$ orbits and $b_1,\dots,b_l$ are in pairwise distinct orbits.}
\label{fig:pac}
\end{figure*}

\begin{prop}\label{prop:pacc}
Let $\relb$ be a structure with finitely many orbits and 
finitely many
pp-definable binary relations. 
Then a given instance
$\rela$ of CSP$(\relb)$
has the PACC if and only if the algorithm presented
in Figure~\ref{fig:pac} does not reject $\rela$.
\end{prop}
\begin{pf}
Suppose that $\rela$ is an instance of CSP$(\relb)$ that has the PACC.
We have to show that for any element $a$ from $A$
there exists an orbit $O$ of $\relb$ 
such that for any choice of $b \in O$
the arc consistency procedure that is called in the inner loop
of the algorithm  in Figure~\ref{fig:pac} does not reject
the instance $[\rela,\{a\}]$ of CSP$([\relb,\{b\}])$. Because
$\rela$ has the PACC, there exists a homomorphism from $\rela$
to $\pow(\relb)$ such that $h(a)$ is a singleton $\{c\}$.
Let $O$ be the orbit of $c$, and let $b$ be the element from
$O$ that is used by the algorithm.
We know that there exists an automorphism $\alpha$
that maps $c$ to $b$. Clearly, the mapping $h'$ defined by 
$x \mapsto \alpha(h(x))$ is a homomorphism from $\rela$ to
$\pow(\relb)$ such that $h'(a)=\{b\}$ is a singleton.
By Lemma~\ref{lem:fin} the structure $[\relb,\{b\}]$ has
finitely many pp-definable unary relations. Proposition~\ref{prop:acc} then shows that the arc consistency
procedure does not reject the instance $[\rela,\{a\}]$ of
CSP$([\relb,\{b\}])$. 
All the implications in this argument can be reversed,
which shows the statement of the proposition.
\end{pf}


\begin{theorem}\label{thm:pac}
Let $\relb$ be a structure with finitely many orbits and
finitely many pp-definable binary relations,
and suppose that PAC solves CSP$(\relb)$.
Then there exists a quadratic-time and linear-space algorithm
that decides CSP$(\relb)$. Moreover, CSP$(\relb)$ can be decided
in linear time with a linear number of processors.
\end{theorem}
\begin{pf}
Let $\rela$ be an instance of CSP$(\relb)$. 
If the algorithm
in Figure~\ref{fig:pac} rejects $\rela$, then it does not have the PACC and hence $\rela$ does not homomorphically map to $\relb$. If the algorithm
in Figure~\ref{fig:pac} does not rejects $\rela$, then $\rela$ has the PACC. By assumption, PAC solves CSP$(\relb)$, and therefore there exists a homomorphism from $\rela$ to $\relb$.

Because the arc consistency procedure uses linear space, the algorithm
in Figure~\ref{fig:pac} can be implemented in linear space as well.
The arc consistency procedure is called a linear number of times (recall that $\relb$ is fixed and not part of the input). Because 
the arc consistency procedure can be implemented such that it uses linear time, the overall running time on a sequential machine is
quadratic in the worst case. However, note that each application
of the arc consistency procedure can be performed on a different
processor.  
\end{pf}

\section{Algebraic Characterization}
\label{sect:alg-characterization}
In this section we present a general algebraic characterization of
those constraint languages where PAC decides CSP$(\relb)$
for an arbitrary  finite or infinite structure $\relb$.

We use the notation $\ind(\pow(\relb)^n)$ to denote the
induced substructure of $\pow(\relb)^n$ whose universe contains
an $n$-tuple of $\pow(\relb)^n$ if and only if
at least one coordinate of the tuple is a singleton.

\begin{theorem}
\label{thm:alg-characterization}
Let $\relb$ be a structure.
PAC decides $\csp(\relb)$ if and only if
for all $n$ there is a homomorphism from all finite substructures
of $\ind(\pow(\relb)^n)$ to $\relb$.
\end{theorem}

\begin{pf}
$(\Leftarrow)$: Suppose that $(\rela, \relb)$ has the PACC.
Then, by definition of the PACC,
 for all $a \in A$, there is a homomorphism $h_a$ from
$[\rela, a]$ to $[\pow(\relb), \{ \{ b \} ~|~ b \in B \}]$.
Let $n = |A|$.  
Now consider the homomorphism $h$ from $\rela$ to
$\pow(\relb)^n$ defined by
$h(x) = \Pi_{a \in A} h_a(x)$.
Notice that for every $a \in A$, the element $h_a(a)$ of the tuple
$h(a)$ is a singleton, and hence $h$ is in fact
a homomorphism from $\rela$ to $\ind(\pow(\relb)^n)$.
Let $\relc$ be the structure that is induced by the image of
$h$ in $\ind(\pow(\relb)^n)$.
Since $\relc$ is finite, it is by assumption homomorphic to $\relb$, 
and by composing homomorphisms we obtain
that there is a homomorphism from $\rela$ to $\relb$.

$(\Rightarrow)$: Let $n \geq 1$, and let
$\relc$ be a finite substructure of $\ind(\pow(\relb)^n)$.
We have to show that $\relc$, viewed as an instance of CSP$(\relb)$, has the PACC, which suffices by assumption.
Let $a$ be any element of the universe of $\relc$.
By definition of $\relc$, we have that $a$ is an $n$-tuple
such that some coordinate, say the $i$th coordinate, is a singleton.  
The projection function $\pi_i$ is a homomorphism from
$[\relc, \{ a \}]$ to 
$[\pow(\relb), \{ \{ b \} ~|~ b \in B \}]$.
\end{pf}

\section{Robustness}
In this section, we demonstrate that
the class of structures $\relb$ such that PAC decides $\csp(\relb)$
is robust in that it satisfies certain closure properties.

We first investigate expansion by pp-definable relations.
Say that a structure $\relb'$ (over $\sigma'$) is a \emph{pp-expansion} of
$\relb$ (over $\sigma$)
if $\relb'$ is an expansion of $\relb$ and for every
symbol $R \in \sigma' \setminus \sigma$, 
it holds that $R^{\relb'}$ is pp-definable over $\relb$.

\begin{theorem}
\label{thm:pp-expansion}
Suppose that PAC decides $\csp(\relb)$.
Then for any pp-expansion $\relb'$ of $\relb$, it holds that
PAC decides $\csp(\relb')$.
\end{theorem}

\begin{pf}
It suffices to show that the theorem holds for an expansion of $\relb$
by (1) an intersection of existing relations, 
(2) a projection of an existing relation,
(3) a product of an existing relation with $B$, or
(4) the equality relation.
In each of these cases, we will consider
an expansion $\relb'$ of $\relb$
where the signature of 
$\relb'$ has an additional symbol $T$.
We will use $\sigma$ to denote the signature of $\relb$,
and so the signature of $\relb'$ will be $\sigma \cup \{ T \}$.

By Theorem~\ref{thm:alg-characterization}, 
it suffices to show that for every $n \geq 1$
and for all finite substructures $\relc'$
of $\ind(\pow(\relb')^n)$ there exists a homomorphism $h$ 
from $\relc'$ to $\relb'$.
Let $C$ be a finite subset of the universe of
$\ind(\pow(\relb)^n)$, let
$\relc$ be the induced substructure of
$\ind(\pow(\relb)^n)$
with universe $C$,
and let
$\relc'$ be the induced substructure of 
$\ind(\pow(\relb')^n)$
with universe $C$.
By Theorem~\ref{thm:alg-characterization},
there is a homomorphism $h$ from $\relc$ to $\relb$.
Since $\relb'$ is an expansion of $\relb$ with just one additional
symbol $T$, it suffices to show 
that $h(T^{\relc'}) \subseteq T^{\relb'}$. 

(1):
Suppose that
$T^{\relb'} = R^{\relb} \cap S^{\relb}$ for $R, S \in \sigma$.
It follows that
$T^{\pow(\relb')} \subseteq R^{\pow(\relb)} \cap S^{\pow(\relb)}$,
from which we obtain
$T^{\relc'} \subseteq 
R^{\relc} \cap 
S^{\relc}$.
For any tuple $\tup{t} \in T^{\relc'}$,
we thus have $h(\tup{t}) \in R^{\relb} \cap S^{\relb}$,
and hence $h(\tup{t}) \in T^{\relb'}$.

(2): 
Suppose that
$T^{\relb'}$ is the relation
$$\{ (t_1, \ldots, t_k) ~|~ \exists t_{k+1}, \ldots, t_{k+l} \mbox{ so that } (t_1, \ldots, t_{k+l}) \in R^{\relb} \}$$
that is, $T^{\relb'}$ is the projection of $R^{\relb}$ onto the first $k$
coordinates; we denote the arity of $R \in \sigma$ by $k+l$.
(We assume that the projection is onto an initial segment 
$\{ 1, \ldots, k \}$ of coordinates
for the sake of notation; a similar argument holds for an arbitrary set
of coordinates.)
We have that $T^{\pow(\relb')}$ is the projection of $R^{\pow(\relb)}$
onto the first $k$ coordinates. Thus, for any tuple 
$\tup{t} \in T^{\relc'}$,
we have that $h(\tup{t})$ is the projection of a tuple in $R^{\relb}$,
and is hence in $T^{\relb'}$.

(3): Suppose that
$T^{\relb'} = R^{\relb} \times B$ for $R \in \sigma$.
Let $\tup{t} = (t_1, \ldots, t_{k+1})$ be any tuple in 
$T^{\relc'}$.
We have that
$(t_1, \ldots, t_k) \in R^{\relc}$,
and hence
$h(t_1, \ldots, t_k) \in R^{\relb}$.
Since we have $h(t_{k+1}) \in B$, it follows that
$h(t_1, \ldots, t_{k+1}) \in T^{\relb'}$.

(4): Suppose that $T^{\relb'} = \{ (b, b) ~|~ b \in B \}$.
For any tuple $(t_1, t_2) \in T^{\pow(\relb')}$, we have
$t_1 = t_2$. For any tuple
$(t_1, t_2) \in T^{\relc'}$
we thus also have $t_1 = t_2$, and
we have $h(t_1, t_2) \in T^{\relb'}$.
\end{pf}

We now consider homomorphic equivalence.  

\begin{theorem}
\label{thm:hom-equivalence}
Let $\relb$ be a structure.
Suppose that PAC decides $\csp(\relb)$ and that
$\relb'$ is a structure that is homomorphically equivalent to $\relb$,
that is, $\relb \rightarrow \relb'$ and $\relb' \rightarrow \relb$.
Then PAC decides $\csp(\relb')$.
\end{theorem}

We first establish the following lemma.

\begin{lemma}
\label{lemma:hom-to-pow}
Let $f$ be a homomorphism from $\relb'$ to $\relb$.
The map $f'$ defined on $\pow(B') \setminus \{ \emptyset \}$
by $f'(U) = \{ f(u) ~|~ u \in U \}$
is a homomorphism from $\pow(\relb')$ to $\pow(\relb)$.
\end{lemma}

\begin{pf}
Let $R$ be a symbol, and let $\tup{t}$ be a tuple in $R^{\pow(\relb')}$.
We have $\tup{t} = (\pi_1 S, \ldots, \pi_k S)$ 
where $S \subseteq R^{\relb}$ and $S \neq \emptyset$.
Define $S' = \{ f(\tup{s}) ~|~ \tup{s} \in S \}$.
We have $S' \subseteq R^{\relb'}$.
As $f'(\tup{t}) = (\pi_1 S', \ldots, \pi_k S')$,
the conclusion follows.
\end{pf}

\begin{pf} (Theorem~\ref{thm:hom-equivalence})
Suppose that $(\rela, \relb')$ has the PACC.
We want to show that $\rela \rightarrow \relb'$.

We first show that $(\rela, \relb)$ has the PACC.
Let $a$ be an element of $A$.
There exists a homomorphism $h$ from $\rela$ to $\pow(\relb')$
such that $h(a)$ is a singleton.
The mapping $f'$ given by 
Lemma~\ref{lemma:hom-to-pow} is a homomorphism from
$\pow(\relb')$ to $\pow(\relb)$ that maps singletons to singletons.
Hence, the map $a \rightarrow f'(h(a))$ is a homomorphism from
$\rela$ to $\pow(\relb)$ mapping $a$ to a singleton.
We thus have that $(\rela, \relb)$ has the PACC.

Since PAC decides $\csp(\relb)$, there is a homomorphism from
$\rela$ to $\relb$.  By hypothesis, there is a homomorphism from
$\relb$ to $\relb'$, and so we obtain that $\rela$ is homomorphic to $\relb$.
\end{pf}

\section{Tractability by PAC}

\paragraph{Slice-semilattice operations.}  We first study
a class of ternary operations.
Recall that a \emph{semilattice operation} is a binary operation that is
associative, commutative, and idempotent,
and that a semilattice operation $\oplus$ is well-defined on finite sets,
that is, for a finite set $S = \{ s_1, \ldots, s_n \}$ we may define
$\oplus(S) =  
\oplus(\oplus(\ldots \oplus(\oplus(s_1, s_2), s_3), \ldots), s_n)$.
We say that a ternary operation $t: B^3 \rightarrow B$ 
is a \emph{slice-semilattice} operation if for every element
$b \in B$, the binary operation $\oplus_b$ defined by
$\oplus_b(x, y) = t(x, y, b)$ is a semilattice operation.
These ternary operations have been studied in~\cite{slaac}; there, 
the following examples were presented.

\begin{example}
Let $B$ be a set, and let $d: B^3 \rightarrow B$
be the operation such that $d(x, y, z)$ is equal to $x$ if
$x = y$, and $z$ otherwise.  This operation is known as the
\emph{dual discriminator} on $B$, and is an example of a slice-semilattice operation. For examples of constraint languages
that have a dual discriminator polymorphism, see e.g.~\cite{JonssonKuivinenNordh}.

\end{example}

\newcommand{\median}{\mathsf{median}}

\begin{example}
Let $B$ be a subset of the rational numbers, and let
$\median: B^3 \rightarrow B$ be the ternary operation on
$B$ that returns the median of its arguments.
(Precisely, given three arguments $x_1$, $x_2$, and $x_3$
in ascending order so that $x_1 \leq x_2 \leq x_3$, the $\median$
operation returns $x_2$.)
This operation is an example of a slice-semilattice operation.
\end{example}

\begin{theorem}
\label{thm:slice-semilattice}
Let $\relb$ be a finite structure that has a slice-semilattice polymorphism.
Then, the problem $\csp(\relb)$ is tractable by PAC.
\end{theorem}

\begin{pf}
Let $f$ denote the slice-semilattice polymorphism.
By Theorem~\ref{thm:alg-characterization}, 
it suffices to show that for every finite substructure $\relc$
of $\ind(\pow(\relb)^n)$
there is a homomorphism $h$ from $\relc$ to $\relb$.

For each element $(S_1, \ldots, S_n)$ in $C$
we define $h(S_1, \ldots, S_n)$ as follows.
Let $g$ be the maximum index such that $S_g$ is a singleton;
we are guaranteed the existence of such an index by the definition of
$\ind(\pow(\relb)^n)$.
We define a sequence of values $b_g, \ldots, b_n \in B$ inductively.
Set $b_g$ to be the value such that $\{ b_g \} = S_g$.
For $i$ with $g < i \leq n$, define $b_i = \oplus_{b_{i-1}} S_i$.
We define $h(S_1, \ldots, S_n) = b_n$.

We now show that $h$ is in fact a homomorphism from 
$\relc$ to $\relb$.
Let $R$ be any symbol of arity $k$.
Suppose that
$((S_1^1, \ldots, S_n^1), \ldots, (S_1^k, \ldots, S_n^k)) \in 
R^{\ind(\pow(\relb)^n)}$.
We define a sequence of tuples $\tup{t_1}, \ldots, \tup{t_n} \in R^{\relb}$
in the following way.
Let $\tup{t_1}$ be any tuple such that
$\tup{t_1} \in (S_1^1 \times \cdots \times S_1^k) \cap R^{\relb}$.
For $i$ with $1 < i \leq n$, we define
$\tup{t_i} = 
(\oplus_{t_{(i-1) 1}} S_i^1, \ldots,
 \oplus_{t_{(i-1) k}} S_i^k).$
Given that $\tup{t_{i-1}}$ is in $R^{\relb}$, we prove that $\tup{t_i}$
is in $R^{\relb}$.  
Let $C_i \subseteq R^{\relb}$ be a set of tuples such that
$(\pi_1(C_i), \ldots, \pi_k(C_i)) = (S_i^1, \ldots, S_i^k)$.
Let $\tup{c^1}, \ldots, \tup{c^m}$ with $m \geq 2$ be a sequence of tuples
such that $\{ \tup{c^1}, \ldots, \tup{c^m} \} = C_i$.
We have 
$\tup{t_i} = 
f(\tup{c^m}, \ldots f(\tup{c^3}, 
f(\tup{c^2}, \tup{c^1}, \tup{t_{i-1}}), \tup{t_{i-1}}) \ldots, \tup{t_{i-1}})$.
Since $f$ is a polymorphism of $R^{\relb}$, we obtain 
$\tup{t_i} \in R^{\relb}$.

Observe now that for each tuple
$(S_1^j, \ldots, S_n^j)$, the values $b_g, \ldots, b_n$ that were computed
to determine 
$h(S_1^j, \ldots, S_n^j) = b_n$ have the property that
for each $i$ with $g \leq i \leq n$,
$b_i = t_{ij}$.  It follows that $h$ is the desired homomorphism.
\end{pf}

It is well-known that the problem 2-SAT can be identified with the problem
$\csp(\relb)$ for the structure $\relb$ with universe $B = \{ 0, 1 \}$
and relations
\begin{center}
$R_{(0, 0)}^{\relb} = \{ 0, 1 \}^2 \setminus \{ (0, 0) \}$

$R_{(0, 1)}^{\relb} = \{ 0, 1 \}^2 \setminus \{ (0, 1) \}$

$R_{(1, 1)}^{\relb} = \{ 0, 1 \}^2 \setminus \{ (1, 1) \}$
\end{center}
It is known, and straightforward to verify, that the dual discriminator
operation on $\{ 0, 1 \}$ is a polymorphism of this structure $\relb$.
We therefore obtain the following.

\begin{theorem}
The problem 2-SAT is tractable by PAC.
\end{theorem}

Let $\sigma$ be the signature $\{ E \}$ where $E$ is a symbol having
arity $2$.  We call a structure $\relg$ over $\sigma$
an \emph{undirected bipartite graph} if $E^{\relg}$ is a symmetric relation,
the universe $G$ of $\relg$ is finite, and $G$ can be viewed
as the disjoint union of two sets $V_0$ and $V_1$ such that
$E^{\relg} \subseteq (V_0 \times V_1) \cup (V_1 \times V_0)$.

\begin{theorem}
Let $\relg$ be an undirected bipartite graph.  
The problem $\csp(\relg)$ is tractable by PAC.
\end{theorem}

\begin{pf}
Let $\relg'$ be the bipartite graph with universe $\{ 0, 1 \}$
and where $E^{\relg'} = \{ (0, 1), (1, 0) \}$.
As $E^{\relg'}$ is pp-definable over the structure $\relb$
corresponding to 2-SAT above, by
$\phi(v_1, v_2) \equiv R_{(0,1)}(v_1, v_2) \wedge R_{(0,1)}(v_2, v_1)$,
we have that PAC decides $\csp(\relg')$ by 
Theorem~\ref{thm:pp-expansion}.

If $E^{\relg}$ is empty, the claim is trivial, so assume that
$(s, s') \in E^{\relg}$.
We claim that $\relg$ and $\relg'$ are homomorphically equivalent, 
which suffices by Theorem~\ref{thm:hom-equivalence}.
The map taking $0 \rightarrow s$ and $1 \rightarrow s'$ is
a homomorphism from $\relg'$ to $\relg$.
The map taking all elements in $V_0$ to $0$ and
all elements in $V_1$ to $1$ is a homomorphism from $\relg$ to $\relg'$.
\end{pf}


We call a finite structure $\reld$ over signature $\{A\}$ where $A$ is a binary relation symbol an \emph{orientation of a cycle} if $D$ can be enumerated as $d_1,\dots,d_n$ 
such that $A^\reld$ contains either $(d_i,d_{i+1})$ or
$(d_{i+1},d_i)$ for all $1 \leq i < n$, contains either 
$(d_n,d_1)$ or $(d_n,d_1)$, and contains no other pairs.
The orientation of a cycle is called \emph{unbalanced} if 
the number of elements $A^\reld$ of the form $(d_i,d_{i+1})$ or $(d_n,d_1)$
is distinct from $n / 2$.
It has been shown in~\cite{FederCycles} that for every unbalanced orientation of a cycle $\reld$ there is a linear order on $D$ such that
$\reld$ is preserved by the median operation with respect to this linear order. 

We therefore have the following result.
\begin{theorem}
Let $\reld$ be an unbalanced orientation of a cycle. Then
CSP$(\reld)$ is tractable via PAC.
\end{theorem}


\paragraph{The Point Algebra in Temporal Reasoning.}
The structure $(\mathbb Q,\leq,\neq)$ is known as the \emph{point algebra} in temporal reasoning. The problem CSP$(\mathbb Q,\leq,\neq)$ can be solved by the path-consistency procedure~\cite{vanBeekCohen}. 

\begin{theorem} 
\label{thm:pa}
CSP$(\mathbb Q,\leq,\neq)$ is tractable via PAC.
\end{theorem}

\begin{pf}
Clearly, the structure $(\mathbb Q;\leq,\neq)$ has only one orbit.
It is well-known that it is also $\omega$-categorical~\cite{Hodges},
and therefore has in particular a finite number of pp-definable binary relations.
To apply Theorem~\ref{thm:pac}, we only have to verify that
PAC decides CSP$(\mathbb Q;\leq,\neq)$. 

Let $\rela$ be an instance of CSP$(\mathbb Q; \leq,\neq)$.
We claim that if there is
a sequence $a_1,\dots,a_k \in A$ 
such that $(a_i,a_{i+1}) \in \; \leq^A$ for all $1 \leq i < k$, $(a_k,a_1) \in \; \leq^\rela$, and $(a_p,a_q) \in \; \neq^\rela$ for some $p,q \in \{1,\dots,k\}$, then there is no homomorphism from $\rela$ to $\pow(\relb)$ such that $h(a_1)$ is a singleton $\{b_1\}$.
Suppose otherwise that there is such a homomorphism $h$. By the definition of $\pow(\relb)$
 there must be a sequence $b_1,\dots,b_k$ such that $b_i \in h(a_i)$ for all $1 \leq i \leq k$ and
 $(b_i,b_{i+1}) \in \; \leq^{\relb}$ for all $1 \leq i < n$. Moreover,
 $(b_k,b_1) \in \; \leq^{\relb}$, and hence $b_1=\dots=b_k$. But then we have $(h(a_p),h(a_q))=(b_1,b_1)Ê\in \; \neq^\relb$, a contradiction.
Hence, the structure $\rela$ does not have the PACC if
$\rela$ has such a sequence $a_1,\dots,a_k$. 
It is known~\cite{vanBeekCohen} that if $\rela$ does not contain such a sequence, then $\rela \rightarrow (\mathbb Q;\leq,\neq)$. This
 shows that PAC decides CSP$(\mathbb Q;\leq,\neq)$.
\end{pf}

\paragraph{Set Constraints.}
Reasoning about sets is one of the most fundamental reasoning
tasks. A tractable set constraint language has been introduced
in~\cite{DrakengrenJonssonSets}. The constraint relations
in this language are containment $X \subseteq Y$ (`every element of $X$ is contained in $Y$'), disjointness $X || Y$ (`$X$ and $Y$ do not have common elements'),
and disequality $X \neq Y$ (`$X$ and $Y$ are distinct'). 
In the CSP for this constraint language we are given a set of
constraints and a set of containment, disjointness, and disequality 
constraints between variables, and we want to know whether it
is possible to assign \emph{sets} (we can without loss of generality assume that we are looking for subsets of the natural numbers; note that we allow the empty set) to these variables such that all the given constraints are satisfied. 
It was shown in~\cite{qe} that this problem can be modeled as 
CSP$((D;\subseteq,||,\neq))$, where $D \subset 2^{\mathbb N}$ is a countably infinite set of subsets of $\mathbb N$, and such that $(D;\subseteq,||,\neq)$ is $\omega$-categorical and has just two
orbits (the orbit for $\emptyset$, and the orbit for all other points).

\begin{theorem}\label{thm:sc}
CSP$((D; \subseteq,||,\neq))$ is tractable via PAC.
\end{theorem}
\begin{pf}
Because $(D; \subseteq,||,\neq)$ is $\omega$-categorical,
it suffices as in the proof of Theorem~\ref{thm:pa} to verify that
PAC decides CSP$((D; \subseteq,||,\neq))$ in order to apply
Theorem~\ref{thm:pac}.

Let $\rela$ be an instance of CSP$((D; \subseteq,||,\neq))$.
We claim that if there are four sequences 
$(a^1_1,\dots,a^1_{k_1})$, \dots, $(a^4_1,\dots,a^4_{k_4})$ of elements 
from $A$ such that
\begin{itemize}
\item $a^1_{k_1}=a^2_{k_2}=a^3_1=a^4_1$,
\item $(a^j_i,a^j_{i+1}) \in \; \subseteq^A$ for all $1 \leq j \leq 4$, $1 \leq i < k_j$,
\item $(a^1_1,a^2_1) \in \neq^\rela$, and
\item $(a^3_{k_3},a^4_{k_4}) \in {||}^{\rela}$.
\end{itemize}
then there is no homomorphism 
$h$ from $\rela$ to $\pow(\relb)$ 
such that $h(a^1_1)$ is a singleton $\{b^1_1\}$. 
Suppose otherwise that there is such a homomorphism $h$.
By the definition of $\pow(\relb)$
 there must be sequences of elements $(b^1_1,\dots,b^1_{k_1})$, \dots,
 $(b^4_1,\dots,b^4_{k_4})$  such that 
 \begin{itemize}
 \item $b^j_i \in h(a^j_i)$ for all $1 \leq j \leq 4$, $1 \leq i \leq k_j$, 
 \item $b^1_{k_1}=b^2_{k_2}=b^3_1=b^4_1$,
\item  $(b^j_i,b^j_{i+1}) \in \; \subseteq^{\relb}$ for all $1 \leq j \leq 4$, $1 \leq i < k_j$, 
\item $(b^1_1,b^2_1) \in \neq^\relb$, and
\item $(b^3_{k_3},b^4_{k_4}) \in {||}^{\relb}$.
 \end{itemize}
The third item and the fourth item together imply that 
$b^1_{k_1}=b^2_{k_2} \neq \emptyset$ (any set that contains two distinct sets cannot be empty).
 The third item and the fifth item together imply that 
 $b^1_{k_1}=b^2_{k_2} = \emptyset$ (any set that is contained in two disjoint subsets must be the empty set), a contradiction.
 
It follows from Lemma 3.7. in~\cite{DrakengrenJonssonSets} 
that if $\rela$ does not contain such sequences, then $\rela \rightarrow (D; \subseteq,||,\neq)$. This shows that PAC decides CSP$(D; \subseteq,||,\neq)$.
\end{pf}

PAC tractability results can also be shown for the basic binary relations
in the spatial reasoning formalism of RCC-5~\cite{RCC5JD}, which is closely related to set constraints, but also for
other known tractable spatial constraint satisfaction problems in qualitative spatial reasoning, e.g., in~\cite{Congruence}.

\small
\bibliography{local}
\bibliographystyle{abbrv}

\end{document}